\documentclass[11pt]{article}

\usepackage[preprint]{acl}

\usepackage{times}
\usepackage{latexsym}

\usepackage{times}
\usepackage{latexsym}
\usepackage{bbm}
\usepackage{booktabs}
\usepackage{makecell}
\usepackage[T1]{fontenc}
\usepackage[utf8]{inputenc}

\usepackage{microtype}

\usepackage{inconsolata}

\usepackage{graphicx}
\usepackage{algorithm}
\usepackage{algpseudocode}
\usepackage{amsmath}

\DeclareRobustCommand{\tool}{%
  \ifmmode
    \text{tool}% in math: same as \text{...}
  \else
    \textnormal{tool}% in text: normal body font
  \fi
}
\usepackage[T1]{fontenc}
\usepackage[utf8]{inputenc}

\usepackage{microtype}

\usepackage{inconsolata}

\usepackage{graphicx}
\usepackage{times}
\usepackage{latexsym}
\usepackage{amsmath}
\usepackage{amsfonts}
\usepackage{url}
\usepackage{graphicx}
\usepackage{booktabs}
\usepackage{xcolor}
\usepackage{rotating}
\usepackage{pifont}
\usepackage{multirow}
\usepackage{enumitem}
\usepackage[table]{xcolor}
\usepackage{subcaption}
\usepackage{tikz}
\usetikzlibrary{shapes.geometric, arrows, positioning}
\usepackage{hyperref}

\usepackage{tcolorbox}
\tcbuselibrary{listings,breakable}
\usepackage{listings}

\usepackage{pdfpages}

\newtcolorbox{promptbox}{
  colback=gray!10,
  colframe=gray!60,
  boxrule=0.3pt,
  arc=2pt,
  left=6pt,
  right=6pt,
  top=6pt,
  bottom=6pt,
  breakable
}

\definecolor{lightpurple}{RGB}{245,240,255}

\newtcolorbox{promptbox1}{
  colback=lightpurple,
  colframe=purple!50!black,
  boxrule=0.4pt,
  arc=3pt,
  left=6pt,
  right=6pt,
  top=6pt,
  bottom=6pt,
  breakable
}

\definecolor{lightgreenbg}{RGB}{235,250,240}

\newtcolorbox{statebox}{
  colback=lightgreenbg,
  colframe=green!50!black,
  boxrule=0.4pt,
  arc=3pt,
  left=6pt,
  right=6pt,
  top=6pt,
  bottom=6pt,
  breakable
}

\definecolor{lightredbg}{RGB}{255,240,240}

\newtcolorbox{judgebox}{
  colback=lightredbg,
  colframe=red!60!black,
  boxrule=0.4pt,
  arc=3pt,
  left=6pt,
  right=6pt,
  top=6pt,
  bottom=6pt,
  breakable
}

\definecolor{lightcyanbg}{RGB}{235,255,255}

\newtcolorbox{toolbox}{
  colback=lightcyanbg,
  colframe=cyan!60!black,
  boxrule=0.4pt,
  arc=3pt,
  left=6pt,
  right=6pt,
  top=6pt,
  bottom=6pt,
  breakable
}

\definecolor{bar_red}{HTML}{C65B5B}   
\definecolor{bar_teal}{HTML}{7FAEA8}

\title{Ada-RS: Adaptive Rejection Sampling for Selective Thinking}

\author{
 \textbf{Yirou Ge},
 \textbf{ Yixi Li},
 \textbf{Alec Chiu},
 \textbf{Shivani Shekhar},
\\
 \textbf{Zijie Pan},
 \textbf{Avinash Thangali},
 \textbf{Yun-Shiuan Chuang},
 \textbf{Chaitanya Kulkarni},
\\
 \textbf{Uma Kona},
 \textbf{Linsey Pang},
 \textbf{Prakhar Mehrotra}
\\
PayPal AI
}

\begin{document}
\maketitle
\begin{abstract}
Large language models (LLMs) are increasingly being deployed in cost- and latency-sensitive settings. While chain-of-thought improves reasoning, it can waste tokens on simple requests. We study selective thinking for tool-using LLMs and introduce Adaptive Rejection Sampling (Ada-RS), an algorithm-agnostic sample filtering framework for learning selective and efficient reasoning. For each given context, Ada-RS scores multiple sampled completions with an adaptive length-penalized reward then applies stochastic rejection sampling to retain only high-reward candidates (or preference pairs) for downstream optimization. We demonstrate how Ada-RS plugs into both preference pair (\textit{e.g.} DPO) or grouped policy optimization strategies (\textit{e.g.} DAPO). Using Qwen3-8B with LoRA on a synthetic tool call-oriented e-commerce benchmark, Ada-RS improves the accuracy–efficiency frontier over standard algorithms by reducing average output tokens by up to $\sim$80\% and reducing thinking rate by up to $\sim$95\% while maintaining or improving tool call accuracy. These results highlight that training-signal selection is a powerful lever for efficient reasoning in latency-sensitive deployments.
\end{abstract}

\section{Introduction}
\label{sec:intro}

\begin{figure*}[t] 
\centering
\includegraphics[width=0.9\linewidth]{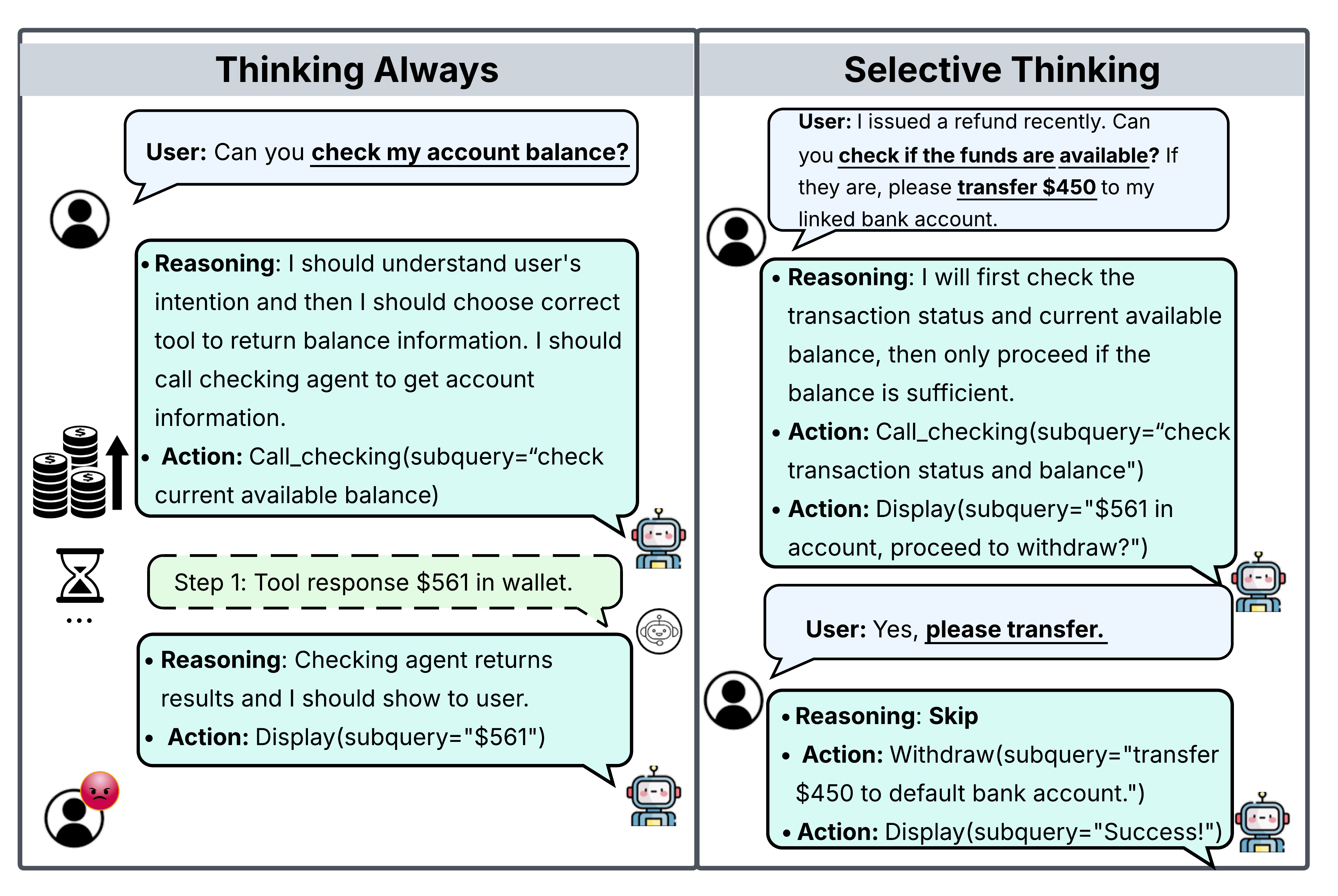}
\vspace{-4mm}
\caption{\textbf{Comparison between thinking always reasoning and selective reasoning in a tool-calling LLM agent.} The agent performs explicit reasoning even for a simple user query, resulting in unnecessary inference cost and latency (left). The agent selectively skips reasoning and directly calls the appropriate tool (right).}
% \vspace{-4mm}
\label{fig:motivation}
% \vspace{-4mm}
\end{figure*}

Large language models (LLMs) are increasingly deployed inside cost- and latency-sensitive systems that facilitate human interactions such as customer service assistants and e-commerce copilots that must respond within tight service-level agreements while handling a large volume of queries. To navigate complex requests, LLMs often rely on explicit chain-of-thought (CoT) \cite{wei2022chain} style reasoning to ensure high quality; however, generating long reasoning traces can often introduce substantial overhead and degrade the user experience, especially when many requests are routine or can be easily handled with short responses (\textit{e.g.} small talk, quick clarifications). As a result, a key practical question in real deployments is not whether models can reason and break down complex tasks, but how to allocate reasoning budget only when it aids in resolving a user's request.

Recent \textit{selective thinking} work has begun to tackle this matter by training or prompting models to modulate reasoning depth through strategies such as learning when to invoke CoT \cite{lou2025adacotparetooptimaladaptivechainofthought}, pruning or compressing reasoning \cite{xiang2025just, fang2025thinklessllmlearnsthink, hou2025thinkprunepruninglongchainofthought}, or balancing accuracy against token usage via reward shaping \cite{yang2025thinkneedselfadaptivechainofthought} and multi-objective optimization \cite{xiang2026selfguided}. These approaches make important progress, but they often rely on boundary tuning or specialized training objectives.

In this work, we study selective thinking through the lens of model training. However, our key design choice is to target a different and complementary lever from existing work: the candidate samples used for learning. We introduce Adaptive Rejection Sampling (Ada-RS), an algorithm-agnostic sampling mechanism that applies rejection sampling over candidates, probabilistically retaining responses that are most informative for learning selective and efficient thinking while downsampling uninformative or unnecessarily verbose samples. Due to its algorithm-agnostic nature, Ada-RS can complement several tuning algorithms such as direct preference optimization (DPO) \cite{rafailov2024directpreferenceoptimizationlanguage} to construct higher-quality preference pairs for preference optimization or group relative policy optimization-style methods (\textit{i.e.,} GRPO-style updates) \cite{shao2024deepseekmathpushinglimitsmathematical, yu2025dapoopensourcellmreinforcement}.

We apply our method to a tool call oriented e-commerce domain setting and find our approach yields favorable accuracy–efficiency trade-offs, substantially reducing token usage by 70\%-80\% without sacrificing performance. These results highlight the importance of how we construct and filter the training signal and how that signal can be leveraged to improve the efficiency of generation, especially for systems where latency and inference cost are first-order product constraints. %\ac{Can we bring this back to latency and how majority of the volume of queries are actually simple ones? For instance, with Oslo, X\% of queries are single intent}

\paragraph{Contributions.}

\begin{enumerate}
    \item We formalize and evaluate \textit{selective thinking} for latency-sensitive systems, emphasizing the practical need to decide \emph{when not to think}.
    \item We propose Ada-RS, an algorithm-agnostic rejection-sampling mechanism for training sample candidates that improves sample efficiency and discourages unnecessary verbosity while preserving long reasoning when useful.
    \item We empirically study different optimization strategies and show that Ada-RS enables stronger accuracy–efficiency trade-offs on tool call oriented tasks in an e-commerce domain setting.
\end{enumerate}

% \begin{figure*}
%     \centering
%     \includegraphics[width=0.8\linewidth]{acl/figures/example_20260203.png}
%     \caption{\textbf{Selective thinking example in a tool calling LLM agent.} For routine requests, the agent skips explicit reasoning and directly issues a minimal tool call to retrieve needed information (left). For more complex or ambiguous requests, the agent produces a reasoning trace and may execute multiple tool calls before returning a final response (right).}
%     \label{fig:motivation}
% \end{figure*}

\section{Related Work}
\label{sec:related}
CoT prompting can substantially improve reasoning \cite{deepseek2025r1, openai2024o1}, but in production settings it often incurs unnecessary latency and cost when long explanations are generated for otherwise simple inputs. Consequently, recent work on efficient reasoning has focused on controlling \emph{when} a model should engage in explicit reasoning (selective thinking) \cite{lou2025adacotparetooptimaladaptivechainofthought, zhang2025adaptthinkreasoningmodelslearn} and \emph{how much} reasoning it should produce, via either (i) explicit, prompt-level control \cite{ma2025reasoningmodelseffectivethinking} or (ii) training-time objectives that encourage selective thinking \cite{sui2025stopoverthinkingsurveyefficient}.

\subsection{Explicitly Controlled Reasoning}
% A straightforward way to reduce overthinking is to \emph{explicitly} control reasoning behavior at inference time, for example with special prompt instructions, formatting constraints, or difficulty-aware prompting. \textit{NoThinking} \cite{ma2025reasoningmodelseffectivethinking} shows that instructing a model to bypass explicit reasoning can preserve accuracy on many low-to-medium complexity tasks while reducing verbose outputs. While effective and easy to deploy, such approaches typically rely on external prompt control and do not necessarily teach the model to \emph{internally} learn when reasoning is warranted; performance can also be sensitive to prompt phrasing and may not generalize across domains \cite{sui2025stopoverthinkingsurveyefficient}. Related work such as AutoThink \cite{autothink} uses prompt-based signals (e.g., an ellipsis prompt) to modulate reasoning depth in R1-style distilled models, and further reinforces this behavior through multi-stage training.

A straightforward way to reduce overthinking is to \emph{explicitly} control reasoning behavior at inference time, for example with special prompt instructions, formatting constraints, or difficulty-aware prompting \cite{ma2025reasoningmodelseffectivethinking, autothink}. While effective and easy to deploy, such approaches typically rely on external prompt control and do not necessarily teach the model to internally learn when reasoning is warranted; performance can also be sensitive to prompt phrasing and may not generalize across domains \cite{sui2025stopoverthinkingsurveyefficient}.

\subsection{Selective Thinking Through Training}
A complementary line of work aims to \emph{learn} selective thinking through training, typically by modifying rewards \cite{yang2025thinkneedselfadaptivechainofthought, lou2025adacotparetooptimaladaptivechainofthought} or objectives \cite{zhang2025adaptthinkreasoningmodelslearn, xiang2026selfguided}, to trade off task success against reasoning cost. Compared to prompt-only control, these training-based methods can yield a single model that better internalizes the decision of when to reason; however, they often require careful tuning of penalty strengths or multi-stage training to avoid degenerate solutions (\textit{e.g.} collapsing to always-think or never-think behavior) \cite{sui2025stopoverthinkingsurveyefficient}.

\subsection{Relation to Our Work}
Our work targets the same goal of selective thinking but focuses on \textbf{training-signal construction and selection}. While prior approaches emphasize reward design or alternative training objectives, we propose \emph{Adaptive Rejection Sampling} (Ada-RS) to stochastically retain the most informative samples (or preference pairs) under an adaptive length penalty. This design aims to reduce the influence of unnecessarily verbose trajectories during training while preserving explicit reasoning on difficult inputs.

\section{Preliminaries}
\label{sec:prelim}

\subsection{Task Overview}

We consider a \emph{tool calling} LLM agent that resolves a user request by optionally invoking e-commerce-related tools (\textit{e.g.}, product search, account information retrieval, transaction look-ups) and then producing a final response. An example of a task in the e-commerce setting we explore is given in Figure~\ref{fig:motivation}. 

At any decision point, the agent observes a context $x$ consisting of the conversation history and any tool
outputs observed so far. Given $x$, the model produces a response
$y = (\texttt{<think>}~t~\texttt{</think>}, a)$ where $t$ is an optional reasoning trace and $a$
is the final answer (including tool calls when applicable). The goal is to learn a policy $\pi_\theta(y\mid x)$ that emits little-to-no reasoning on easy or simple instances while still using reasoning when it materially improves correctness. 
\section{Methods}
\label{sec:methods}

\begin{figure*}[tb!]
    \centering    \includegraphics[width=\linewidth]{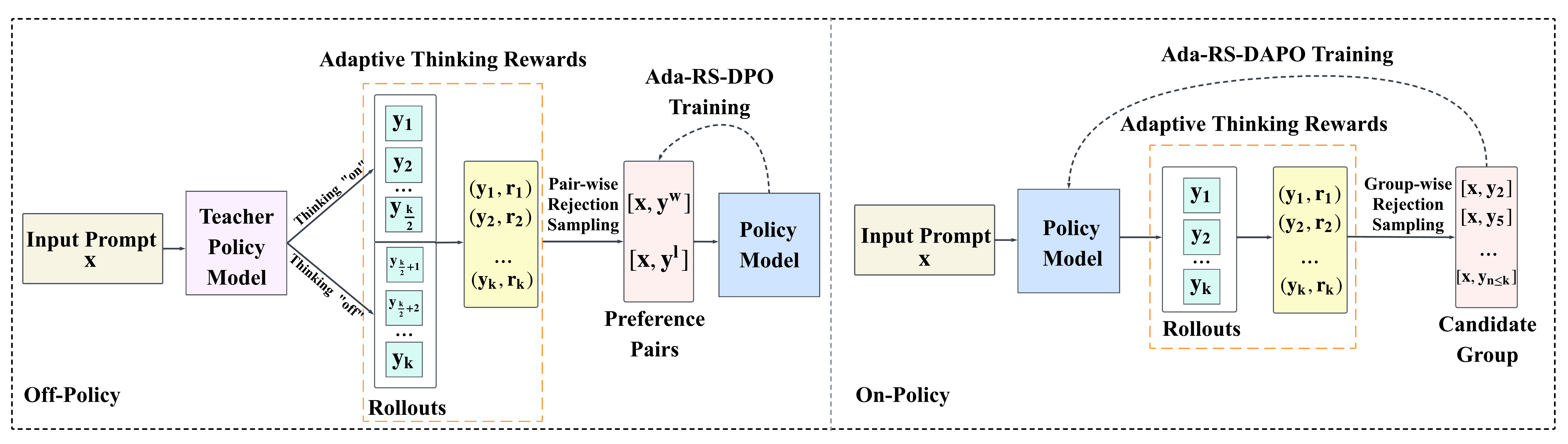}
    \caption{\textbf{Overview of the proposed Ads-RS training framework.} \textbf{\emph{Off-policy Ada-RS-DPO}} training pipeline (left). Given an input context $x$, a teacher policy model generates multiple rollouts. Ada-RS performs pair-wise rejection sampling based on adaptive thinking reward signals for selective thinking to construct high-quality preference pairs, which are then used to optimize the student policy model via DPO training. \textbf{\emph{On-policy Ada-RS-DAPO}} training pipeline (right). The current policy model generates multiple rollouts for the input context. Ada-RS applies group-wise rejection sampling based on adaptive thinking rewards to select informative candidate subsets. The resulting candidate group is used to update the policy through an on-policy DAPO training.}
    \label{fig:pipeline}
\end{figure*}

\subsection{Adaptive Rejection Sampling (Ada-RS)}
Ada-RS is a lightweight \emph{sample selection} mechanism that can be plugged into both off-policy and on-policy training objectives that rely on sampled completions. An overview of the framework and how we apply it are illustrated in Figure~\ref{fig:pipeline}. 

For each context $x$, we draw $K$ training sample candidates $\{y_i\}_{i=1}^K \sim \pi_\phi(\cdot\mid x)$,
assign each candidate an adaptive efficiency-aware reward, and then apply stochastic rejection sampling to retain only the most informative candidates (or candidate pairs) for downstream optimization.

\subsubsection{Adaptive length penalty (ALP)}
Given rollouts $\{y_i\}_{i=1}^K \sim \pi_\theta(\cdot\mid x)$, we define a composite reward that trades off task success and reasoning cost:
\begin{equation}
  r(y_i, x) = \mathbbm{1}(y_i, x) - \alpha \cdot s_K(x) \cdot |t_i|
  \label{eq:ada_rs_reward}
\end{equation}
Here $\mathbbm{1}(y_i, x) \in \{0,1\}$ indicates whether the rollout $y_i$ solves the task for prompt $x$
(\textit{i.e.} correct tool call). $|t_i|$ penalizes the length of the reasoning trace in the rollout (we use the number of sentences inside \texttt{<think>} as a proxy; $|t_i|=0$ when the \texttt{<think>} block is empty). The key adaptive component is
\begin{equation}
  s_K(x) = \frac{1}{K}\sum_{i=1}^{K} \mathbbm{1}(y_i, x)
  \label{eq:solve_rate}
\end{equation}
an online estimate of how easy the prompt is under the current policy. When $s_K(x)$ is high, Ada-RS applies a stronger length penalty, discouraging unnecessary reasoning on easy prompts; when it is low, the penalty shrinks, allowing longer reasoning on harder prompts. This follows the spirit of \cite{xiang2025just}, which scales a length cost by an online solve-rate estimate.

\subsubsection{Rejection sampling over training sample candidates}
Given rewards $\{r_i\}_{i=1}^K$ for context $x$, Ada-RS performs rejection sampling to preferentially retain higher-reward (more correct and/or more efficient) samples while keeping stochasticity for diversity. We support:

\paragraph{Pair-wise rejection sampling (for preference learning).}
For a candidate pair $(i,j)$, define $\Delta_{ij} = r_i - r_j$ and accept the pair with probability
\begin{equation}
p_{ij} \;=\; \exp\!\left(\frac{\Delta_{ij} - \Delta_{\text{max}}}{\beta_{\mathrm{rs}}}\right),
\label{eq:pairwise_rs}
\end{equation}
where $\beta_{\mathrm{rs}}$ is a temperature controlling selectivity hyperparameter and $\Delta_{\text{max}} = \max_{i<j}(\Delta_{ij})$. The accepted pair is converted into a preference example $(x, y^w, y^l)$ by setting $y^w=\arg\max(r_i,r_j)$ and $y^l=\arg\min(r_i,r_j)$. This builds on recent successes found with rejection sampling \cite{liu2023statistical} and utilizing reward gaps for preference pair optimization \cite{khaki2024rsdpohybridrejectionsampling}.

\paragraph{Group-wise rejection sampling (for grouped policy optimization).}
Alternatively, we accept each candidate $y_i$ independently based on its standardized reward within the group:
\begin{equation}
p_i \;=\; \min\left(\exp\!\left(\frac{(r_i-\mu)/\sigma}{\beta_{\mathrm{rs}}}\right),1\right),
\label{eq:groupwise_rs}
\end{equation}
where $\mu$ and $\sigma$ are the mean and standard deviation of $\{r_i\}_{i=1}^K$ for the prompt. Smaller $\beta_{\mathrm{rs}}$ concentrates training on above-mean samples; larger values keep more diverse candidates.
% \sh{DAPO already has a length aware penalty. how do we justify the adaptive length penalty works better? \yx{I also tried DAPO with overlong reward shaping enabled (which penalizes the reward only when the response is overlong), and the results were similar to DAPO, as it still couldn't learn selective thinking. Not sure if it should be included to results since it's an optional optimization implemented by DAPO?}}
% \yg{I think one argument we can give here is that even though the DAPO paper proposes an optional length-aware penalty, it only penalizes the model when the response exceeds a given length (i.e. overlong). This design contains the model's reasoning within a specific length for all prompts and is not difficulty-aware. This is not aligned with our selective thinking objective, which justifies why we went forward with our design.}

\subsubsection{Ada-RS-DPO}
Ada-RS-DPO uses pair-wise rejection sampling to construct higher-quality preference pairs for DPO. For each context $x$, we sample $K$ candidates of equal amounts across thinking-enabled/disabled examples, compute rewards via Eq.~\ref{eq:ada_rs_reward}, and accept candidate pairs with Eq.~\ref{eq:pairwise_rs}. Using the accepted preference pairs, we then optimize a DPO objective that includes an auxiliary negative log-likelihood (NLL) loss term to stabilize learning and preserve language modeling quality as described in \cite{pang2024iterativereasoningpreferenceoptimization}. Psuedocode for this algorithm can be found in Appendix Algorithm~\ref{alg:ada-rs-dpo}.
% \begin{equation}
% \mathcal{L}_{\mathrm{Ada\text{-}RS\text{-}DPO}} \;=\; \mathcal{L}_{\mathrm{DPO}}
% \;+\; \lambda_{\mathrm{NLL}} \cdot \mathcal{L}_{\mathrm{NLL}}(x,y^w),
% \label{eq:ada_rs_dpo_loss}
% \end{equation}
% where $\lambda_{\mathrm{NLL}}$ weights the
% auxiliary term.

\subsubsection{Ada-RS-DAPO}
We also integrate Ada-RS with Decoupled Clip and Dynamic Sampling Policy Optimization (DAPO) \cite{yu2025dapoopensourcellmreinforcement}, which we use as our grouped
policy optimization backbone. For each context $x$, we sample a group of $K$ on-policy candidates $\{y_i\}_{i=1}^K \sim \pi_\theta(\cdot\mid x)$, compute rewards $\{r_i\}^{K}_{i=1}$ using  Eq.~\ref{eq:ada_rs_reward}, and
apply group-wise rejection sampling using the acceptable probability from Eq.~\ref{eq:groupwise_rs} to stochastically filter candidates before the DAPO update. The DAPO objective is then computed on the retained candidates, leaving the underlying DAPO loss unchanged while concentrating gradient updates on trajectories that are both correct and efficiency-favorable. This yields a simple plug-in mechanism to bias grouped on-policy learning toward selective thinking. Psuedocode for this algorithm can be found in Appendix Algorithm~\ref{alg:ada-rs-dapo}.

\section{Experimental Settings}
\label{sec:exp-settings}

All experiments use Qwen3-8B as the reasoning base model \cite{qwen3technicalreport} with a LoRA adapter \cite{hu2022lora} for all training runs. Additional details on training algorithm hyperparameters can be found in Appendix~\ref{sec:training_details} and Appendix~\ref{sec:hyperparameters}. 

\subsection{Evaluation Setting}

We evaluate our methodology on a synthetic multi-turn, multi-step e-commerce dataset designed to mirror common user personas and tasks modeled on general themes observed on e-commerce platforms. The tools available in the dataset mirror those in the $\tau^2$-Bench retail benchmark \cite{barres2025tau2benchevaluatingconversationalagents}. An example can be seen in Figure~\ref{fig:motivation}. Overall, our base training dataset consists of 15,000 tool invocations across 8,026 conversations, which span across 121 tasks and 8 user personas. Our evaluation dataset consists of 2,510 tool calls from 367 conversations, which span across 48 tasks and 8 user personas.

\subsection{Metrics}
The key metrics we evaluate are as follows:
\begin{itemize}
    \item \textbf{Thinking Rate}: the percentage of instances in which the model produces a non-empty reasoning trace. This metric measures how often the model chooses to engage in explicit reasoning.
    \item \textbf{Output Token Length}: the average number of generated output tokens produced by the model (inclusive of reasoning) as a measure of token efficiency. This serves as a deployment-relevant proxy for inference cost and latency.
    \item \textbf{Tool Call Accuracy}: whether the model selects the correct tool and produces the correct arguments, ensuring functional correctness when reasoning is skipped.
\end{itemize}
% \paragraph{Thinking Rate.}
% We define the thinking rate as the fraction of instances in which the model produces a non-empty reasoning trace. This metric measures how often the model chooses to engage in explicit reasoning.
% \paragraph{Output Token Length.}
% We report the average number of generated output tokens produced by the model (inclusive of reasoning) as a measure of token efficiency. This serves as a deployment-relevant proxy for inference cost and latency.
% \paragraph{Tool Call Accuracy.}
% We measure tool call accuracy by evaluating whether the model selects the correct tool and produces the correct arguments, ensuring functional correctness when reasoning is skipped.
\subsection{Baselines and Ablation Studies}
We compare prompt-only baselines, supervised fine-tuning (SFT), DPO, and DAPO as well as ablations and hyperparameter sweeps for Ada-RS.

\subsubsection{Prompt-only and SFT baselines}
We include two \textit{no-fine-tuning} (NFT) baselines that differ only by prompting:
\textit{NFT (Thinking-On)}, where reasoning is always invoked, and \textit{NFT (Thinking-Off)}, where explicit reasoning never occurs (\textit{i.e.} forced empty \texttt{<think>}). An SFT baseline is established by applying SFT on a dataset of 75\% reasoning and 25\% non-reasoning data.

\subsubsection{DPO baselines and ablations}
We evaluate DPO-based training with and without the components used in Ada-RS-DPO: ALP reward, NLL auxiliary loss, and rejection sampling. When not using the ALP reward, we utilize a simple reward strategy to construct preference pairs where correct responses are preferred over incorrect responses and subsequently more concise responses are preferred amongst the correct responses. %\yg{To summarize: For each prompt, we sample multiple rollouts from the current policy across thinking-enabled/disabled modes and evaluate correctness. If both correct and incorrect responses are present, we construct the preference pair by selecting a correct response as chosen and an incorrect response as rejected. If all responses are correct, we instead prefer brevity by selecting the shortest response as chosen and the longest as rejected.}

\subsubsection{DAPO baselines and ablations}
We also evaluate Ada-RS in a grouped policy optimization setting using DAPO as the backbone. We utilize DAPO with only accuracy as a reward function and DAPO with the ALP reward function, both without rejection sampling, as baseline ablations for grouped policy optimization.
\section{Results}
\label{sec:results}

\subsection{Learning When to Reason}
\begin{figure}
    \centering
    \includegraphics[width=\linewidth]{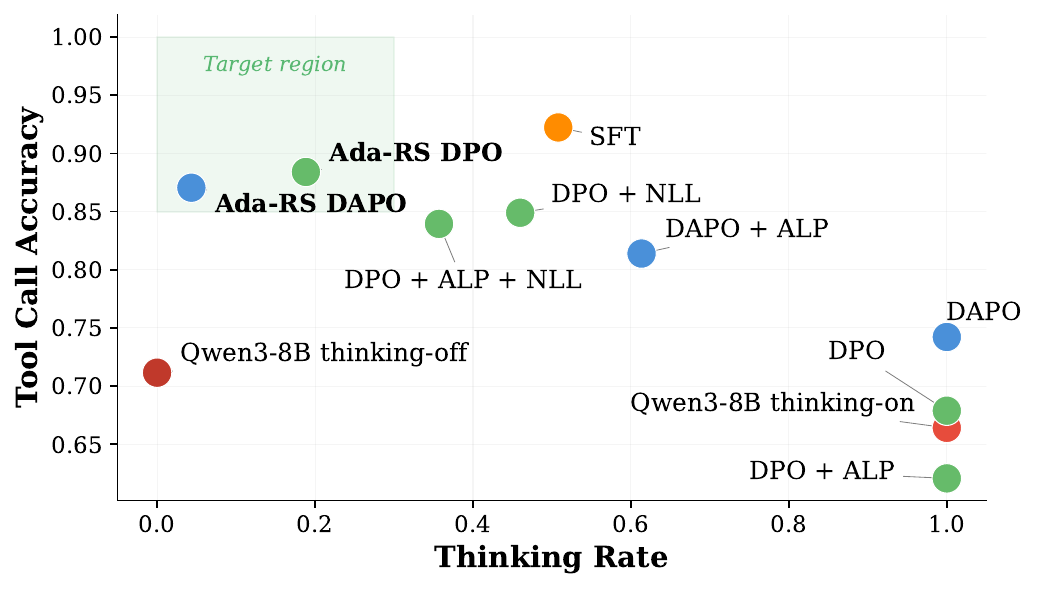}
    \caption{\textbf{Tool call accuracy versus Thinking Rate across methods.} The most favorable target region (high accuracy, low thinking rate) is highlighted. Points are colored by algorithm: DPO (green), DAPO (blue), SFT (orange), and no-fine-tuning/base model (red).}
    \label{fig:accuracy_vs_thinking_rate}
\end{figure}

\begin{figure}
    \centering
    \includegraphics[width=\linewidth]{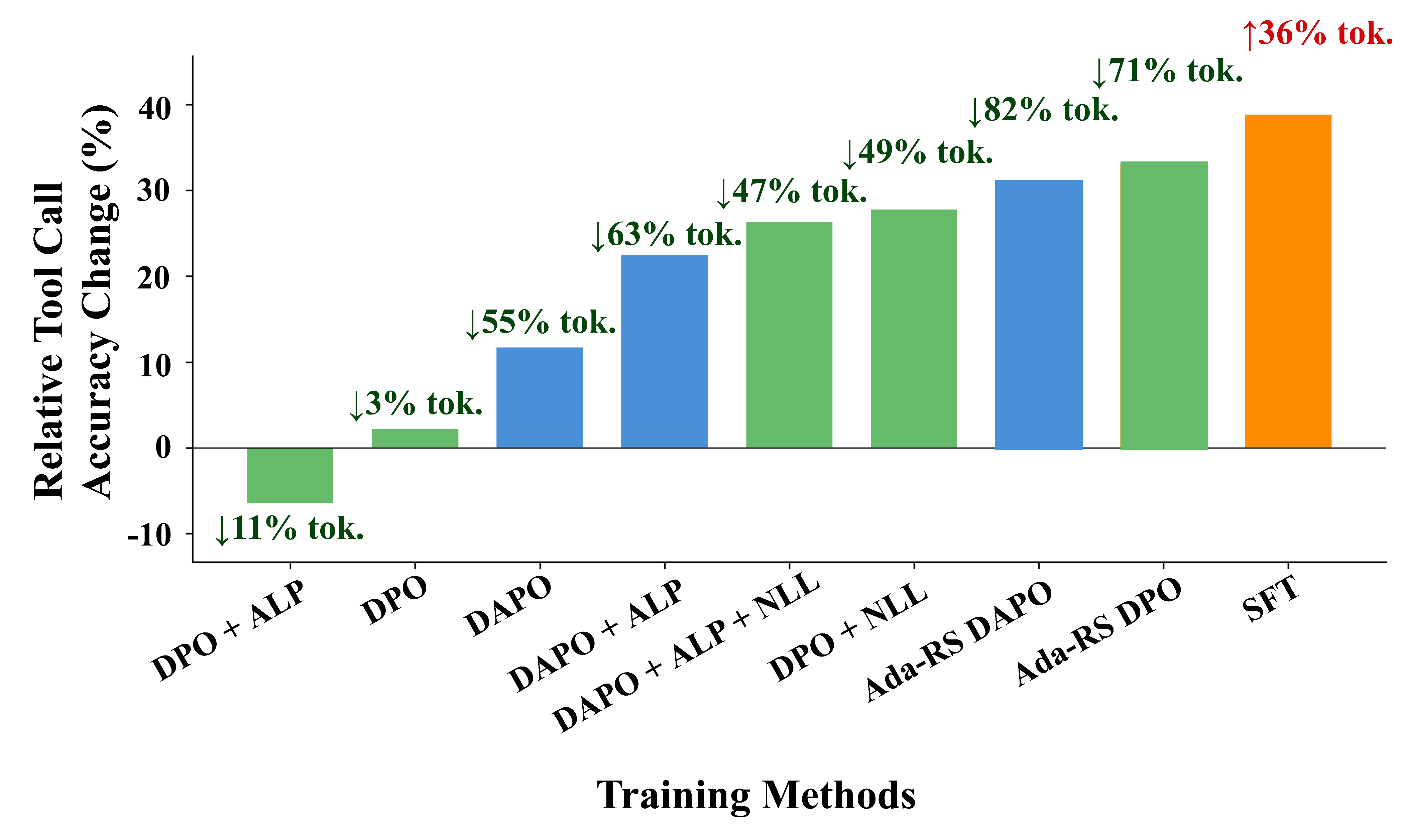}
    \caption{\textbf{Tool call accuracy and average Output token across methods relative to the Qwen3-8B base model with thinking enabled.} Numbers above the bars show percentage decrease in average amount of output tokens relative to the base model. Bars are colored by algorithm: DPO (green), DAPO (blue), and SFT (orange).}
    \label{fig:relative_token_cost}
\end{figure}

We first examine whether different training strategies induce the ability to selectively think. SFT is able to successfully induce thinking to trigger only about half the time (Figure~\ref{fig:accuracy_vs_thinking_rate}) but with high verbosity (Figure~\ref{fig:relative_token_cost}). In contrast, DPO applied directly to the base model with a simple preference pair generation strategy fails to remove the model's default \textit{always-think} behavior at all, suggesting that preference optimization alone does not reliably teach the model \emph{when} to deliberate (Figure~\ref{fig:accuracy_vs_thinking_rate}).

Ada-RS changes this behavior by filtering training sample candidates using an efficiency-aware reward. Ada-RS drastically reduces how often explicit reasoning is invoked overall (Figure~\ref{fig:accuracy_vs_thinking_rate}), while still allocating concise reasoning budgets to maintain accuracy (Figure~\ref{fig:relative_token_cost}).

\subsection{Accuracy-Efficiency Trade-offs}
We can first focus on the two best extremes in terms of accuracy and generated output tokens. While SFT attains the highest overall tool call accuracy, it does so with the largest average output length (Figure~\ref{fig:relative_token_cost}), reflecting high inference cost in latency-sensitive deployments. Meanwhile, the extreme of \textit{NFT Thinking-Off} produces short outputs and never triggers reasoning but yields substantially lower tool call accuracy (Figure~\ref{fig:relative_token_cost}), indicating that simply suppressing reasoning at inference time is insufficient for reliability. Applying the Ada-RS framework substantially improves the accuracy-efficiency frontier. Ada-RS-DPO achieves a markedly better operating point: attaining similar tool call accuracy to SFT but with low output tokens and a very low thinking rate (Figure~\ref{fig:relative_token_cost}). Ada-RS-DAPO further improves the frontier: reaching similar accuracy and token cost (Figure~\ref{fig:relative_token_cost}) as Ada-RS-DPO while reducing the thinking rate even further (Figure~\ref{fig:accuracy_vs_thinking_rate}). Together, these results suggest Ada-RS is complementary to both pairwise preference optimization and grouped policy optimization.
% Compared to several combinations of DPO, Ada-RS further reduces the thinking rate (Figure~\ref{fig:accuracy_vs_thinking_rate}) at an essentially identical token cost (Figure~\ref{fig:relative_token_cost}), indicating that rejection sampling amplifies selective thinking behavior beyond what is obtained by the reward/auxiliary objective alone or in combination. 
\subsection{Ablation Studies}
To further understand the effect of different components in Ada-RS, we ablate the specific components that drive selective thinking and efficiency. We further provide the hyperparameter analysis in Appendix~\ref{sec:hyperparameters}.
\begin{table}[h]
\centering
\caption{\textbf{Ablation study results for the effect of using NLL loss with DPO loss.} For both experiments, $\beta_{rs}=0.1$ and $\alpha = 0.01$. RS denotes the use of the rejection sampling procedure used in Ada-RS for preference pairs.}
\label{tab:nll_ablation}
\small
\begin{tabular}{lccc}
\hline
\textbf{Method} & \makecell{\textbf{Accuracy}\\\textbf{(\%)}} & \makecell{\textbf{Avg Output}\\\textbf{Tokens}} & \makecell{\textbf{Thinking}\\\textbf{Rate (\%)}} \\
\hline
DPO ALP + RS & 63.82  & 451.64 & 100.00 \\
Ada-RS-DPO   & 89.24  & 87.81  & 6.10   \\
\hline
\end{tabular}
\end{table}

% \begin{table}[]
% \caption{\textbf{Ablation study results for the effect of using NLL loss with DPO loss.} For both experiments, $\beta=0.1$ and $\alpha = 0.01$. RS denotes the use of the rejection sampling procedure used in Ada-RS for preference pairs.}
% \label{tab:nll_ablation}
% \resizebox{\columnwidth}{!}{
% \begin{tabular}{llll}
% \hline
%                                                   & \textbf{Accuracy (\%)} & \textbf{Avg Output Tokens} & \textbf{Thinking Rate (\%)} \\
%                                                   \hline
% DPO ALP + RS & 63.82  & 451.64 & 100.00      \\
% Ada-RS-DPO & 89.24 & 87.81           & 6.10 \\
% \hline
% \end{tabular}
% }
% \end{table}
\paragraph{Rejection sampling without stabilization fails to learn selective thinking.}
Applying rejection sampling on top of the ALP reward without the auxiliary NLL stabilization term leads to degenerate behavior (poor accuracy and an always-think policy) (Table~\ref{tab:nll_ablation}). This highlights that naive filtering alone can destabilize learning and that stabilization is important for maintaining correctness while optimizing efficiency.

\paragraph{NLL induces selectivity; ALP improves efficiency; Ada-RS amplifies.}
DPO alone fails to induce selective thinking (Figure~\ref{fig:accuracy_vs_thinking_rate}). Adding the NLL term yields large improvements in selective thinking and accuracy (Figure~\ref{fig:accuracy_vs_thinking_rate}), while the ALP reward improves token efficiency by discouraging verbosity (especially on prompts that have a high solve rate under the current policy) (Figure~\ref{fig:relative_token_cost}). The addition of rejection sampling from Ada-RS then further amplifies this effect beyond what is obtained by the reward and auxiliary objectives by concentrating updates on high-reward (correct and efficient) trajectories (Figure~\ref{fig:accuracy_vs_thinking_rate},~\ref{fig:relative_token_cost}).

\section{Limitations and Future Work}

Our study has several limitations that point to promising directions for future work. First, our experiments focus on a single domain and model size conducted using an internal simulated environment that reflects a specific interaction structure and task distribution in the e-commerce domain. While this setting enables controlled study of selective reasoning behavior under specific system constraints, extending the analysis to additional domains and model scales would help to further assess generality. Second, our evaluation emphasizes \emph{per-step} tool call accuracy, which may not fully capture end-to-end task success in multi-turn settings (\textit{e.g.} whether a user goal is ultimately satisfied). Future work should include goal-completion metrics and other user-facing outcomes.

\section{Conclusion}

We studied selective thinking for latency- and cost-sensitive LLM deployments, where the practical objective is not simply to reason well, but to allocate explicit reasoning only when it materially improves tool behavior. To this end, we introduced \textbf{Adaptive Rejection Sampling (Ada-RS)}, an algorithm-agnostic mechanism that filters training sample candidates using an adaptive efficiency-aware reward, downweighting unnecessarily verbose trajectories while preserving reasoning on harder inputs. Across tuning backbones, Ada-RS consistently improved the accuracy-efficiency frontier in our e-commerce tool call setting: it reduced overall output length and the frequency of explicit reasoning while maintaining strong tool call accuracy. Overall, these findings highlight that \emph{how} we construct and filter training signal can be a first-order lever for deploying reasoning-capable models under strict product constraints, and that selective thinking can be induced without relying on inference-time gating or prompt switches.

\bibliography{custom}

\appendix

\newpage
\appendix
\onecolumn

\section{Additional Training Details}
\label{sec:training_details}
All algorithms initialize from base Qwen3-8B model weights and are optimized using a standard Adam-based optimization. In all experiments, we used the same LoRA configuration: $r=32$, $\alpha = 32$, and $\text{dropout}=0.05$ with LoRA applied to the target modules $q$, $k$, $v$, and $o$. 

For SFT, we used a learning rate of $5 \times 10^{-4}$, while the DPO experiments employ $5 \times 10^{-5}$, and DAPO experiments use $5 \times 10^{-6}$. A warm-up ratio of 0.03 is applied in all three setups. Different learning rates are selected for each method due to their varying sensitivities to learning dynamics. Extensive hyperparameter tuning was not performed for any of the individual approaches.

In experiments involving a combination of DPO and NLL loss, we set $\lambda_{\text{NLL}} = 1$. For both DPO and DAPO experiments utilizing rollouts, the temperature is set to 1 for generation, with $K = 6$ for DPO and $K = 8$ for DAPO. A smaller $K$ is chosen for DPO due to its pairwise design, which would otherwise result in an exponentially larger number of pairs.

% \begin{figure*}
%     \centering
%     \includegraphics[width=\linewidth]{acl/figures/difficulty_breakdown.png}
%     \caption{\textbf{Pareto frontiers for accuracy-efficiency (measured by averaged output tokens) trade-offs across difficulty tiers.} Ada-RS-DPO uses $\beta_{rs}=0.01$ while Ada-RS-DAPO uses $\beta_{rs}=0.5$. Both Ada-RS methods use $\alpha=0.005$.}
%     \label{fig:difficulty_breakdown}
% \end{figure*}

% \begin{figure}
%     \centering
%     \includegraphics[width=\linewidth]{acl/figures/pareto.png}
%     \caption{Caption}
%     \label{fig:pareto}
% \end{figure}

% \begin{figure}
%     \centering
%     \includegraphics[width=\linewidth]{acl/figures/accuracy_hyperparameter.png}
%     \caption{Caption}
%     \label{fig:accuracy_hyperparameter}
% \end{figure}

\begin{table*}[t]
\caption{\textbf{Hyperparameter sweep for $\boldsymbol{\beta_{rs}}$ and $\boldsymbol{\alpha}$}.  Sample numbers are reported to the nearest thousand. RS Acceptance Rate denotes the empirical acceptance rate from rejection sampling. Training time decrease denotes the fold change relative to using all produced samples (120k samples or preference pairs).}
\label{tab:hyperparameter_sweep}
\centering
\small
\begin{tabular}{lccccccccc}
\hline
\textbf{Method} &
  $\boldsymbol{\beta_{rs}}$ &
  $\boldsymbol{\alpha}$ &
  \textbf{Acc.} &
  \makecell{\textbf{Avg Output}\\\textbf{Tokens}} &
  \makecell{\textbf{Avg Reason.}\\\textbf{Tokens}} &
  \makecell{\textbf{Thinking}\\\textbf{Rate}} &
  \makecell{\textbf{Training}\\\textbf{Samples}} &
  \makecell{\textbf{RS Accept.}\\\textbf{Rate}} &
  \makecell{\textbf{Train Time}\\\textbf{Decrease}} \\
\hline
Ada-RS-DPO  & 1       & 0.01  & 84.10 & 236.52 & 126.54 & 50.60 & 96k  & 79.00 & 0.79x \\
Ada-RS-DPO  & 0.5     & 0.01  & 86.26 & 233.06 & 114.25 & 50.48 & 86k  & 71.00 & 0.71x \\
Ada-RS-DPO  & 0.1     & 0.01  & 89.24 & 87.81  & 0.94   & 6.10  & 53k  & 43.00 & 0.43x \\

Ada-RS-DPO  & 0.1     & 0.001  & 87.69 & 88.00  & 3.77   & 28.64  & 53k  & 43.00 & 0.43x \\

Ada-RS-DPO  & 0.01    & 0.01  & 89.68 & 84.81  & 0.83   & 16.97 & 28k  & 23.00 & 0.23x \\
Ada-RS-DPO  & $\approx$0 & 0.01  & 90.04 & 87.21 & 0.87   & 13.03 & 12k  & 10.00 & 0.1x  \\
\hline
Ada-RS-DAPO & 1       & 0.005 & 86.33 & 152.25 & 75.32  & 6.69  & 115k & 96.50 & 1.3x  \\
Ada-RS-DAPO & 0.5     & 0.005 & 87.05 & 81.90  & 36.31  & 4.34  & 78k  & 65.20 & 0.7x  \\
Ada-RS-DAPO & 0.1     & 0.005 & 87.97 & 81.66  & 37.68  & 4.22  & 54k  & 45.50 & 0.5x \\
Ada-RS-DAPO & 0.1     & 0.001 & 73.39 & 118.64  & 41.55  & 93.98  & 54k  & 45.50 & 0.5x \\
\hline
\end{tabular}
\end{table*}

\section{Hyperparameter Sweeps}
\label{sec:hyperparameters}

We present our full hyperparameter sweeps results in Table~\ref{tab:hyperparameter_sweep}. We first fix $\alpha$ and a hyperparameter sweep over $\beta_{rs}$. For Ada-RS-DPO, we observe a moderate range of $\beta_{rs}$ (0 - 0.1) can provide the best trade-off between accuracy and selective reasoning. Larger $\beta_{rs}$ values lead to excessive thinking and lower the tool calling accuracy. Similar trend is observed for Ada-RS-DAPO, a slightly higher range of $\beta_{rs}$ (0.1 - 0.5) better supports selecting thinking, yielding more consistent improvement. 

We further examine the effect of decreasing $\alpha$. Since $\alpha$ controls the strength of the length pressure; weaker penalties increase the frequency of thinking and can increase total output length. We observe in both experiment groups that overly small $\alpha$ significantly degrade performance and impair the model’s adaptive thinking capability, often resulting in unstable or excessive reasoning behaviors.

\begin{algorithm*}[h]
\caption{Ada-RS-DPO (Off-Policy)}
\label{alg:ada-rs-dpo}
\begin{algorithmic}[1]
\Require Dataset $\mathcal{D}$ of contexts $x$; teacher policy $\pi_{\phi}$; student policy $\pi_{\theta}$;
rollout count $K$; ALP weight $\alpha$; RS temperature $\beta_{\mathrm{rs}}$; NLL weight $\lambda_{\mathrm{NLL}}$.
\State $\mathcal{P}\leftarrow \emptyset$ \Comment{preference pairs for DPO}
\For{each context $x \in \mathcal{D}$}
\State \textbf{Generate mixed candidates (thinking-on/off):}
\State Sample $\{y_i^{\text{on}}\}_{i=1}^{K/2} \sim \pi_{\phi}(\cdot \mid x,\ \texttt{think}=1)$
\State Sample $\{y_i^{\text{off}}\}_{i=1}^{K/2} \sim \pi_{\phi}(\cdot \mid x,\ \texttt{think}=0)$
\State $\{y_i\}_{i=1}^{K} \leftarrow \{y^{\text{on}}\}\cup\{y^{\text{off}}\}$
\State Compute correctness $c_i \leftarrow \mathbbm{1}(y_i, x)\in\{0,1\}$ for each $y_i$
\State Extract think trace $t_i$ from $y_i$ and set $\ell_i \leftarrow |t_i|$ \Comment{$|t_i|=0$ if empty \texttt{<think>}}
\State $s_K(x)\leftarrow \frac{1}{K}\sum_{i=1}^{K} c_i$ \Comment{solve-rate estimate, Eq.~(\ref{eq:solve_rate})}
\State $r_i \leftarrow c_i - \alpha\cdot s_K(x)\cdot \ell_i$ \Comment{ALP reward, Eq.~(\ref{eq:ada_rs_reward})}
\State \textbf{Pair-wise rejection sampling to build preferences:}
\State $\Delta_{\max} \leftarrow \max_{i<j} (r_i - r_j)$
\For{each unordered pair $(i,j)$ with $i<j$}
  \State $\Delta_{ij}\leftarrow r_i-r_j$
  \State $p_{ij} \leftarrow \exp\!\left((\Delta_{ij}-\Delta_{\max})/\beta_{\mathrm{rs}}\right)$ \Comment{Eq.~(\ref{eq:pairwise_rs})}
  \State Draw $u \sim \mathrm{Uniform}(0,1)$
  \If{$u < p_{ij}$}
    \State $y_w \leftarrow \arg\max\{r_i,r_j\}$;\quad $y_\ell \leftarrow \arg\min\{r_i,r_j\}$
    \State $\mathcal{P} \leftarrow \mathcal{P}\cup \{(x, y_w, y_\ell)\}$
    \EndIf
\EndFor
\EndFor
\State Update $\theta$ with DPO on $\mathcal{P}$ plus auxiliary NLL:
\State \hspace{1.2em} minimize $\ \mathcal{L}_{\mathrm{DPO}}(\mathcal{P};\theta)\;+\;\lambda_{\mathrm{NLL}}\mathcal{L}_{\mathrm{NLL}}(\mathcal{P};\theta)$
\end{algorithmic}
\end{algorithm*}

\begin{algorithm*}[t]
\caption{Ada-RS-DAPO}
\label{alg:ada-rs-dapo}
\begin{algorithmic}[1]
\Require Dataset $\mathcal{D}$ of contexts $x$; current policy $\pi_{\theta}$;
rollout count $K$; ALP weight $\alpha$; RS temperature $\beta_{\mathrm{rs}}$.
\For{each training step}
  \State Sample minibatch $\{x_b\}_{b=1}^B \sim \mathcal{D}$
  \For{each context $x$ in minibatch}
    \State \textbf{On-policy rollout group:}
    \State Sample $\{y_i\}_{i=1}^{K} \sim \pi_{\theta}(\cdot \mid x)$
    \State Compute correctness $c_i \leftarrow \mathbbm{1}(y_i, x)$ and think-length $\ell_i \leftarrow |t_i|$ for each $y_i$
    \State $s_K(x)\leftarrow \frac{1}{K}\sum_{i=1}^{K} c_i$ \Comment{solve-rate estimate, Eq.~(\ref{eq:solve_rate})}
    \State $r_i \leftarrow c_i - \alpha\cdot s_K(x)\cdot \ell_i$ \Comment{ALP reward, Eq.~(\ref{eq:ada_rs_reward})}
    \State \textbf{Group-wise rejection sampling:}
    \State $\mu \leftarrow \frac{1}{K}\sum_{i=1}^{K} r_i$;\quad $\sigma \leftarrow \sqrt{\frac{1}{K}\sum_{i=1}^{K}(r_i-\mu)^2}$
    \State $\mathcal{Y}' \leftarrow \emptyset$ \Comment{retained candidate group}
    \For{each candidate $y_i$}
      \State $p_i \leftarrow \min\Big(\exp\big(((r_i-\mu)/\sigma)/\beta_{\mathrm{rs}}\big),\ 1\Big)$ \Comment{Eq.~(\ref{eq:groupwise_rs})}
      \State Draw $u \sim \mathrm{Uniform}(0,1)$
      \If{$u < p_i$}
        \State $\mathcal{Y}' \leftarrow \mathcal{Y}' \cup \{(y_i, r_i)\}$
      \EndIf
    \EndFor
    \State \textbf{DAPO update on retained group:}
    \State Apply one DAPO optimization step using $(x,\mathcal{Y}')$
  \EndFor
\EndFor
\end{algorithmic}
\end{algorithm*}

\end{document}